\documentclass[11p]{article}
\usepackage{oldlfont,rawfonts}

\usepackage{times}
\usepackage{graphicx}
\usepackage{latexsym}
\usepackage{color}

\newcommand{\FOMOL}{\mbox{${\mathcal{D}}al$ }}
\newcommand{\DYS}{\mbox{$\MM = ({\mathcal W}, \{{\mathcal{S}}_{w}: w \in  {\mathcal{W}}\}, {\mathcal{A}}, {\bf R},(\tau_{0}, \tau _{1}, \tau_2,  \tau_3))$ }}
\newcommand{\WW}{\mbox{$\mathcal{W}$ }}
\newcommand{\MM}{\mbox{$\mathcal{M}$ }}
\newcommand{\Sf}{{\bf{S4\,}}}
\newenvironment{proof}[1]{Proof #1}{{\sc q.e.d.}}

\newcommand{\bi}{\begin{itemize}}
\newcommand{\ei}{\end{itemize}}
\newcommand{\be}{\begin{enumerate}}
\newcommand{\ee}{\end{enumerate}}
\newcommand{\bax}{\begin{ax}}
\newcommand{\eax}{\end{ax}}

 \newcommand{\hide}[1]{}

\newcommand{\true}{\mbox{$\top$}}
\newcommand{\false}{\mbox{$\bot$}}

\newcommand{\doppel}[1]{#1\hspace{-0.25cm}#1}
\newcommand{\Card}[1]{\mbox{$\mid \hspace{-0.1cm}#1\hspace{-0.1cm}\mid$}}

\def\PREREQ{\mbox{PREREQ}}
\newcommand{\prereq}[1]{\ifmmode\PREREQ(#1)\else$\PREREQ(#1)$\fi}

\def\JUST{\mbox{JUST}}
\newcommand{\just}[1]{\ifmmode\JUST(#1)\else$\JUST(#1)$\fi}

\def\CONS{\mbox{CONS}}
\newcommand{\cons}[1]{\ifmmode\CONS(#1)\else$\CONS(#1)$\fi}

\newcommand{\Xstar}[1]{\mbox{$K^\star$}}

\newcommand{\Xnset}[2]{\mbox{$\{#1_1,#1_2,\ldots ,#1_#2\} $}}
\newcommand{\Xnseq}[2]{\mbox{$#1_1,#1_2,\ldots ,#1_{#2}$}}

\newcommand{\Xnterm}[3]{\mbox{$#1(#2_1,#2_2,\ldots ,#2_{#3}) $}}

\newcommand{\Xnforl}[3]{\mbox{$\vdash_{#3} \neg #1_1 \vee \ldots
\vee \neg #1_#2$}}

 \newtheorem{de}{Definition}
 \newtheorem{theo}{Theorem}
 
 \newtheorem{lem}{Lemma}
 
 \newtheorem{re}{Remark}
\newtheorem{ax}{Axiom}

\newtheorem{ex}{Example }

\title{A first-order Temporal Logic for Actions}

\author{
Camilla Schwind\\
 {\small\it LIF- CNRS, Luminy} \\
 {\small\it Universit\'e de la M\'editerran\'ee} \\
{\small\it 163 Avenue de Luminy, Case 901}
{\small\it 13288 Marseille Cedex 9, France, }\\
{\small\it E-mail: schwind@lif.univ-mrs.fr} }

\date{}

\begin{document}

\maketitle

\thispagestyle{empty}


\begin{abstract}
We present a  multi-modal action logic with  first-order
modalities, which contain  terms  which can be unified with the
terms inside the subsequent formulas and  which can be quantified.
This makes it possible   to handle simultaneously  time and
states. We discuss applications of this  language to action theory
where it is possible to express many temporal aspects of actions,
as for example, beginning,  end,  time points, delayed
preconditions and results, duration and many others. We present
tableaux rules for a decidable fragment of this logic.
\end{abstract}

\section{Introduction}

\begin{sloppypar}

Most action theories consider actions being specified by their
preconditions and their results.   The temporal structure of an
action system is then defined by the sequence of actions that
occur. A world is conceived as a graph of situations where every
link from one node to the next node is considered as an action
transition. This yields also a temporal structure of the action
space, namely sequences of actions can be considered defining
sequences of world states. The action occurs instantantly at one
moment and its results are true at the ``next'' moment.

However, the temporal structure of actions can be much more
complex and  complicated.
\begin{itemize}
\item Actions may have a duration.

 \item The  results may be true    before the action is completed
 or after it is finished.

 \item Actions may have preconditions which have to have been true during
 some interval preceding the action occurrence.

 \end{itemize}

 In order to represent complex temporal structures, underlying
 actions' occurrences, we have developed an action logic which
 allows to handle both states and time simultanuously.

 We want to be able to express, for instance that action $a$ occurs
 at moment $t$ if conditions $p_1$, \dots $p_n$ have been true
 during the
 intervals $i_1$, \dots all preceding $t$.

Here we present an approach where  it is possible to represent
actions in a complex temporal environment. In reality, actions
have sometimes a beginning time, a duration, preconditions which
may also have temporal aspects ; and the results may be true only
instances after the end of the action performance. For an example,
consider the action of calling an elevator, taking place at
instance $t_1$. Depending on the actual situation this action may
cause the elevator to move only many instances later, to stop
still later, an so on. The action of pressing the button of a
traffic light, in order to get green light to traverse the street
may result in a switch immediately or  after some seconds and
 in another switch after some more minutes.

In order to represent such issues, we define a modal action logic,
where the modalities are terms containing variables which can be
quantified. The same variables can occur inside the modalities as
well as in the formulas after the modalities, allowing for
unification between action term components and logical terms. This
language makes it possible to express reasoning on states and the
action terms allow to express temporal aspects of the actions.

\section{The first-order modal action logic \FOMOL}

The language of first-order action logic, $\mathcal L$ is an
extension of the language of classical predicate logic, ${\mathcal
L}_0$. ${\mathcal L}_0$ consists of
  a set  of  variables $x, y, x_1, y_1, \ldots$,
   a set $\doppel{F}$ of  function
symbols $F$, where  $\Card{F} \in \omega$ is the    arity of $F$,
   a set $\doppel{P}$ of  predicate
symbols $P$,  including $\top$ and $\bot$, where  $\Card{P} \in \omega$ is the    arity of $P$,
  an equality symbol $=$,
   the logical symbols $\neg$, $\wedge$ ,$\forall$.
Terms and formulas are defined as usual and so are $\exists$ and
$\vee$. We denote by $V_t$ the set  of all terms of ${\mathcal
L}_0$. If $\phi$ is a formula and $x$ a  variable then we say that
x is bounded in  $\phi$, when it occurs in a subformula $\forall x
\phi$. $x$ occurs free in $\phi$  if it occurs in $\phi$  and is
not bounded in $\phi$.

\paragraph*{Action terms}

The language for action operators consists of \bi \item
 a set $\doppel{A}$ of {\em action symbols}  $a_1, a_2, \dots $ where
$\Card{a} \in \omega$ is the arity of $a$ and such that $\doppel{A} \cap \doppel{P} = \emptyset$
\end{itemize}
Action terms are built  from action symbols and terms of ${\mathcal L}_0$.
\begin{itemize}
\item if $a$ is an action symbol of arity $n$ and $t_1, \dots t_{n}$ are  terms of
${\mathcal L}_0$, then $a(t_1, \dots t_{n}) $ is an action term.
\end{itemize}
An action term is called {\em grounded } if no  variable occurs free in it.
The set of grounded action terms is denoted by $\doppel{A}_{t}$.\\

\paragraph*{Action operators}

If $a$ and $a_1, a_2, \dots a_n$ are action terms, then
\begin{itemize}
\item  $[a]$ is an action operator
\item $[a_1;a_2; \dots a_n]$ is an action operator
\end{itemize}

\paragraph*{Modal operator}  
$\Box$ is the standard modal operator ( \Sf)

\noindent For $n=0$, the corresponding action  operator is noted $[\varepsilon]$.\\
An action operator is called {\em grounded } if all the action terms occurring in it are grounded.\\
Example: $[a]$, $[a(c_1, c_2, c_3)]$  are grounded, $[a(x, c_2,
y)]$ is not grounded.

\paragraph*{Action formulas}
\begin{itemize}
\item  If $\phi$ is a first-order formula and $[A]$ is an action operator, then $[A]\phi$ is an action formula.
\item  If $\phi$ is a first-order formula and $\Box$ is the modal operator, then $\Box\phi$ is an action formula.

\item  If $\phi$ is an action formula and $x$ is a  variable, then $\forall x\phi$ is an action formula.

\end{itemize}

\paragraph*{Instantiation}

If $\phi$ is a formula and $t$ is a  term, then  $\phi_{t}^{x}$ is the formula obtained
from $\phi$ by replacing every free occurrence of $x$ by $t$. If $t$ is
the name of an element of a set $\mathcal O$ then  $\phi_{t}^{x}$ is called
$\mathcal O$-instance of $\phi$.

Example:
\[\begin{array}{ll}
& [a(x,c)](\neg\phi(c,x) \vee \psi(x))_{c_1}^{x} = [a(c_1,c)](\neg\phi(c,c_1) \vee \psi(c_1))\\
& [a_1;a_2;a_3(c,y)]P(c,y)_{c_3}^{y} = [a_1;a_2;a_3(c,c_3)]P(c,c_3)
\end{array} \]

\noindent A formula is called {\em grounded} if there is no
variable occurring free in it.

\subsection{Semantical Characterization of \FOMOL}

A \FOMOL structure is defined as a Kripke-type structure, such that the transition relation between worlds  depends on grounded action terms.
\\

\noindent A  \FOMOL structure is a tuple  
$\MM = ({\mathcal W}, 
\{{\mathcal S}_{w}: w \in {\mathcal W}\}, {\mathcal A},  {\bf  R}, \tau )$, where
\begin{itemize}
\item $\mathcal W$ is a set of {\em worlds}
\item for every $w\in \mathcal W$, ${\mathcal S}_{w} = ({\mathcal O}, {\mathcal F}_{w},  {\mathcal P}_{w})$ is a classical structure, where $\mathcal O$ is the set of individual   objects (the same set in all worlds), ${\mathcal F}_{w}$ is a set  of functions over $\mathcal O$ and ${\mathcal P}_{w}$ is a set of predicates over $\mathcal O$.
\item  $\mathcal A$ is a set of {\em action functions}, for $f\in \mathcal A$, 
 $f:{\mathcal W} \times \underbrace{{\mathcal O}  \times \ldots \times {\mathcal O}}_{n} \longrightarrow {2}^{\mathcal W}, n \in \omega$. Action functions will characterize the action operators (every action symbol of arity $n$ in $\doppel{A}$ will be associeted with an action function of arity $n+1$).
\item  $ {\bf  R} \subseteq  {\mathcal W} \times  \mathcal W$ is a binary accessability relation on  $\mathcal W$,  which will characterize the modal operater $\Box$. We will write  $R(w) = \{w' : (w,w')\in R\}$.

\item $\tau$ is a valuation, $ \tau = (\tau_{0}, \tau_{1}, \tau_2,
{\tau}_3)$, where  $\tau_0 $ is a function assigning 
objects from ${\mathcal O}$ to terms. In order to
speak about objects from ${\mathcal O}$, we introduce into the
language, for every $o \in \mathcal O$, an $o$-place function symbol
(denoted equally $o$, for simplicity).

$\tau_1 $ is a function assigning, for every world $w\in \mathcal W$, functions (from $\mathcal F$) to  function symbols (from $\doppel{F}$), of the same arity, \\
$\tau_1:\mathcal{W}\times\doppel{F}  \longrightarrow \mathcal F$ such that $\Card{\tau_1(w,F)} = \Card{F}$.\\
 $\tau_2 $ is a function assigning, for every world $w\in \mathcal W$, predicates  to predicate symbols of the same arities, \\
$\tau_2:\mathcal{W}\times\doppel{P}  \longrightarrow \mathcal P$, such that $\Card{\tau_2(w,P)} = \Card{P}$.\\
$\tau_3 $ is a function assigning action functions to action symbols, \\
$\tau_3:\doppel{A}  \longrightarrow \mathcal A$, such that $\Card{\tau_3(a)} = \Card{a} +1$
\item   $ \tau_3(a)(w, \tau_{0}(t_1), \ldots, \tau_{0}(t_{n})) \subseteq   R(w)$. If a world can be reached from $w$ by the execution of action $a(\Xnseq{t}{n})$ then it is   accessible   (via the relation $R$).

\end{itemize}

$\tau_{0},\tau_{1}, \tau_2$ and ${\tau}_3$ define the valuation $\tau$ as follows:

\begin{itemize}
\item  If $ \Xnterm{F}{t}{m}$ is a term then \\
$\tau_0(w, \Xnterm{F}{t}{m}) = \tau_1(w,F)(\tau_0(w,t_1),\tau_0(w,t_2), \ldots, \tau_0(w,t_m))$.
\item if $P$ is an n-ary predicate symbol and $\Xnseq{t}{n}$ are free object variables then \\
$\tau(w,P\Xnseq{t}{n}) = \true$ iff $( \tau_{0}(t_1), \dots  \tau_{0}(t_n)) \in \tau_{2}(w,P)$
\item $\tau(w, t_1 = t_2) = \true$ iff $\tau(w, t_1) = \tau(w, t_2)$
\item
$\tau(w,\neg \phi) = \true$ iff   $\tau(w, \phi) = \false$
\item
$\tau(w, \phi \wedge \psi) = \true$ iff   $\tau(w, \phi) = \tau(w, \psi) = \true$
\item
$\tau(w,\forall x \phi) = \true$ iff   for every $o\in \mathcal O$
 $\tau(w, \phi_{o}^{x}) = \true$

\item $\tau(w,[a(\Xnseq{t}{n})]\phi) = \true$ iff for every $w' \in \tau_3(a)(w, \tau_{0}(t_1), \ldots,  \tau_{0}(t_{n}))$, \\
$\tau(w',\phi) = \true$
\item $\tau(w,\Box \phi) = \true$ iff for every $w' \in R(w)$, 
$\tau(w',\phi) = \true$
\end{itemize}

Let $\phi $ be a formula and $\Xnseq{x}{n}$ be the free object
variables occurring  in $\phi$. Then $\tau(s,\phi)=t$ iff for
every tuple $\Xnseq{t}{n}$ of ground terms,
$$\tau(w,\phi_{\Xnseq{t}{n}}^{\Xnseq{x}{n}}) = t$$

A formula $\phi$ is called valid in state $w \in \mathcal W$ of a
\FOMOL-structure $\mathcal M$ iff $\tau(w,\phi) = \true$. This is
denoted by  ${\mathcal M}, w \models \phi$.  We also say then that
$\phi $ is satisfiable. A formula  $\phi $ is called valid in a
\FOMOL - structure $\mathcal M$ with the set of states $\mathcal
W$, iff $\phi $ is valid in  every $w \in \mathcal W$. We denote
that by $\mathcal{M} \models \phi$. A formula   $\phi $ is called
\FOMOL - valid iff $\phi$ is valid in every \FOMOL - structure.
This is denoted by $\models_{\FOMOL} \phi $. We suppress the index
\FOMOL, whenever it  is clear from the context, in which system we
are.
\begin{re}\label{rebot}
$[a]\bot $ is satisfiable and we have $\tau(w,[a]\bot) = \true$ iff 
$\tau_3(a)(w,  \tau_{0}(t_1), \ldots,  \tau_{0}(t_{n})) = \emptyset$ \end{re}

\subsection{Axioms and inference rules of \FOMOL}

In addition to the axioms and inference rules of classical first -
order logic and those of the system $K$, which rule the action
operators including $[\varepsilon]$, and those of the system  and
${\mathcal S}_4$, which rule the  operator $\Box$, we have the
following axioms and  inference rules, (where $[A]$, $[A_1]$ and
$[A_2]$ are arbitrary action operators):

\[\begin{array}{llll}
& [A1] & [A_1;A_2] \alpha \leftrightarrow [A_1][A_2] \alpha  & \\

& [A2] & \Box \alpha \rightarrow  [A]\alpha & \\

& [A3] & [\varepsilon] \alpha \rightarrow  \alpha & \\

& [A4] & \forall x \alpha \rightarrow  \alpha_{c}^{x} \mbox{ for any individual term } c \mbox{ of } \mathcal L & \\

& [A5] & \forall x [X] \alpha \leftrightarrow [X] \forall x  \alpha  \mbox{ for any modal operator X, with no occurrence of x }  &  \\
\end{array} \]

\[\begin{array}{llll}
 & [R1] & \mbox{ From } \alpha  & \mbox{ infer } \Box \alpha \\
& [R2] & \mbox{ From } \alpha   \rightarrow \beta & \mbox{ infer } \alpha   \rightarrow   \forall x \beta  \mbox{ provided }x  \mbox{ has no free occurrence in }  \alpha \\

\end{array} \]

$\vdash_{\FOMOL}$ is defined as usual, such that $ \vdash_{\FOMOL}
\phi$ for any  instance $\phi$ of one of the axioms; and   $
\vdash_{\FOMOL} \psi$, whenever $\psi$ can be inferred from
$\phi$, for any $\phi$, such that $\vdash_{\FOMOL} \phi$ by use of
one of the inference rules. Again, we suppress the index \FOMOL,
whenever it is clear from the context, in which system we are.

\subsection{Soundness, Completeness, Decidability} The \FOMOL-logic is
sound and complete:

\begin{theo}  $\vdash_{\FOMOL}\phi$ if and only if $\phi$ is \FOMOL- valid
($\models_{\FOMOL}\phi$) \end{theo}

\noindent The soundness proof is easy and the completeness proof
goes along the lines of  completeness proofs for modal logics  by
construction of a canonical model. The proof, which can be found
in the appendix, bears several modifications according to the
specific language which allows to quantify over terms occurring
within modal operators.\\

\FOMOL is a first order language and therefore undecidable in the
general case. But for action logics, we will make use of decidable
subsets of \FOMOL.

\FOMOL is very close to term modal logic introduced by \cite{FittingThalmannVoronkov:01}. Term modal logic allows terms in general as modalities, whereas our action logic only admits action terms. Moreover \FOMOL contains the  \Sf modal operator $\Box$ which is not part of term modal logic.

\section{Temporal Action Theories}

 Using \FOMOL, we can modelize temporal aspects of dynamic actions. The
 modal logic allows to define  action operators as modalities
 \cite{GiordanoMartelliSchwind:00,Schwind:98}. The first order
 logic is used    to formulate actions at a more general
 level. Here, we show an example where in addition to the relative
 representation of time by the modal operators, it is possible to
 express time points by terms.

We presuppose a time axis, linearly ordered (dense or continuous
or discrete). Given a \FOMOL-structure, we will define a
transitive relation on the set of states, $\mathcal{W}$, which
will be related to the order on $\mathcal{T}$.

\begin{de}
Let  \DYS  ,  be a \FOMOL-model.  Then $w
\prec_0 w'$ iff  $ \exists a \in  {\mathcal A}$ of arity $n$ and there are  terms $ t_1, \dots, t_n$,    such that $w' \in f(w,  t_1, \dots, t_n)$.
Let be  $  \preceq  $ the reflexive and transitive  closure of $\prec_0$. 

\end{de}

Intuitively, this means that $w \prec w'$ if we can possibly
``reach'' $w'$ from $w$ by performing actions  \Xnseq{a}{n}.
Obviously, $\preceq$ is transitive and reflexive. Since we want to
``link'' worlds of $\mathcal W$ to time points in $\mathcal T$,
which is ordered, $\prec$ must also be antisymmetric. The temporal
entrenchment of the states is  defined by a homomorphism
 $time: {\mathcal W} \longrightarrow {\mathcal T}$ from ${\mathcal W}$ into ${\mathcal T}$,
 where $w \preceq  w' $ implies $time(w) \leq time(w')$.
  Using this construction,   actions operators can be defined admitting complex
  temporal structures, including beginning and ending instances and a
  duration, which can be  $0$, when the result is immediate, or  $\Delta\in {\mathcal
  T}$. The   preconditions  and results of actions   can be defined to occur
  at freely determinable time instances before or after the
  instance when the action occurs.
 When an action $a$ occurs in the  state $w$, $time(w)$ gives us the time point at which
  $a$ occurs. If the duration of the action is  $\Delta$, the time point of the
  resulting state $w'$ is $time(w') = time(w) + \Delta$.
\\

\noindent In this particular framework, we define

 \bi

 \item
 {\em Action terms} as binary action predicates
  $a(t,d,\overrightarrow{x})$, where $t$ denotes the instance
   on which $a$ occurs and $d$ denotes the duration of $a$, i.e. the
   interval on ${\mathcal T}$ after which the results of $a$ will hold. ,
    $\overrightarrow{x}$ is the sequence of other variables denoting the other   entities or objects
     involved in the action occurrence.

To give an example, let ${\mathcal T}= \{1, \dots,24\}$ be
discrete and finite,
  denoting the
 hours during one day. Then action $move(t,3,TGV,Marseille, Paris)$ is the action ``train
  TGV goes from  Marseille to Paris, the duration being   $3$ hours''.

\item {\em Action axioms}. An action axiom is characterized by a
precondition $\pi(t, \overrightarrow{x})$ and a result  $\rho(t+d,
\overrightarrow{x})$, where $\pi$ and $\rho$ are \FOMOL formulas
describing  all preconditions and  results of  action $a$.

To continue the previous  example,  the action execution axiom of
the move-action
 is $ at(t,x,y)\rightarrow [move(t,d,x,y,z)]at(t+d,x,z)$ (and can be instantiated to
  $ at(6,TGV,Marseille)\rightarrow [move(6,3,TGV,Marseille,Paris)]at(9,TGV,Paris)$), which
 means:
if $x$ is at $y$ at instance $t$, then, after moving
 from $y$ to $z$, $x$ is at $z$ at instance $t+d$.
\ei

 The general  form of an action law is
$$\pi(t_1,\overrightarrow{x_1 }) \rightarrow [a(t,d,\overrightarrow{x_2})]\rho(t_2,
\overrightarrow{x_3}), {\mbox  where  }~ \overrightarrow{x_1}\cup
\overrightarrow{x_2}
 \subseteq \overrightarrow{x_3}$$

\section{Example}
The following example is due to  Lewis \cite{Lewis:00} and has
been discussed by Halpern and Pearl  in \cite{HalpernPearl:01} in
the framework of a theory of causation. Interestingly, this
example defines actions with a complex temporal structure.

 \textit{Billy and Suzanne throw rocks at a bottle. Suzanne
throws first and
 her rock arrives first. The bottle shatters. When Billy's
rock gets to where the bottle used to be, there is nothing there
but flying shards of glass. Without Suzanne's throw, the impact of
Billy's rock on the intact bottle would have been one of the final
steps in the causal chain from Billy's throw to the shattering of
the bottle. But, thanks to Suzanne's preempting throw, that impact
never happens.}

 In our formulation, we focalize on the temporal
structure of the throw action. We consider that the action occurs
along a continuous (or dense) time axis, $[0,\infty[$. We define
one action term for ``throw'', $T$, and two predicates $H$ for
``hits'' and $BB$ for ``the bottle is broken''. The action term
$T(t,d,p)$ means that ``person $p$ throws a stone to a  bottle at
instance $t$ and the result of the action      (the stone hits its
target) occurs at instance $t+d$''. The formula $H(t,p)$ means
that ``the stone thrown by person $p$ hits the bottle at instance
$t$ and formula $BB(t)$ means that the bottle is broken at
instance $t$. The intended result of the action is to hit the
bottle, but this result can only be achieved if the bottle is
still at the intended place and nothing else has been happened to
it, namely if it is not broken in the meantime. In this example it
is not enough to have the precondition that the bottle is there
and not broken at the instance of throwing, but it must  be
non-broken at the moment when the action is to be completed, just
before it is to be hit. Therefore the action law for ``throw'' has
a precondition which must hold after the instance when the action
occurs.

\begin{ex}\label{ex:stone}
 The following set of laws   represents the framework of
 this story:
\[\begin{array}{llll}
  (1) &  \neg
    BB(t+ d)
\rightarrow [ T(t,d,p)] H(t+ d,p)   \\

(2)& \Box (H(t,p) \rightarrow BB(t+d_1))   \\
(3) & \Box (BB(t) \rightarrow \forall t' (t < t' \rightarrow
BB(t')))\\
(4) & \neg BB(0)
\end{array}\]

\noindent (1) is the action law for successful execution of the
throw action, (2) describes the impact of hitting the bottle
($d_1$ is infinitesimally  small) and the general law (3) says
that a broken bottle remains broken ``forever'' \footnote{In this
example we focus on the temporal relations between the different
instances of throwing (by Suzanne and by Billy), so we neglected
other preconditions, as for example having a stone, heavy enough,
but not too heavy, having members enabling the person  to throw,
seeing the  object to aim, etc. The throw action defined here is
highly abstracted for the purpose of our temporal action theory.}.
\end{ex}

\noindent Several scenarios can happen within   this framework.
Here we discuss the scenario where Suzanne throws at instance $0$
and Billy throws some instance later\footnote{In order to express
  that action $a$occurs, we write $[a]\top$, which simply means
  that action $a$ occurs (even when nothing can be said about its
  results). It is always
possible to throw a stone at a bottle, even if the intended result
of hitting cannot be achieved.}.

\[\begin{array}{ll}
  (5) &    < T(0,d_s,suzy)>\top  \\
  (6) & <T(t_1,d_b,billy)>\top

\end{array}\]

\noindent We need these ``empty results'', because
\\

\noindent Three cases can then be distinguished:

\be

\item The moment when the bottle can be hit (and broken) after
Suzanne's throw ($d_s + d_1$) occurs  \textbf{before}  Billy's
stone could possibly hit the bottle $t_1 + d_b$.

\[\begin{array}{ll}
  (7) &  d_s + d_1  < t_1 + d_b \\
  (8) & \Box (BB( d_s + d_1) \rightarrow BB(t_1 + d_b)) \mbox{  from (3) and (7) }\\
  (9) & \neg BB(d_s)  \mbox{  by persistency from (4)\footnote{We use ``weak'' frame
   laws of the form $f\rightarrow[a]f$ which are added to a scenario whenever $[a]\neg f$ cannot be derived from the scenario's laws. }}\\
  (10) & [ T(0,d_s,suzy)] H( d_s,suzy) \mbox{  from (1) and (9) }\\
  (11) & [T(0,d_s,suzy)]BB(d_s + d_1) \mbox{  from (2), (10), K for the action modality and   (A2) }\\
  (12) & [T(0,d_s,suzy)]BB(t_1 + d_b) \mbox{  from (11), (8), K  and   (A2) }

\end{array}\]

In this scenario, the law $ \neg     BB(t_1+ d_b) \rightarrow [
T(t_1,d_b,billy)] H(t_1+ d_b,billy) $ cannot be used to derive
$T(t_1,d_b,billy)] H(t_1+ d_b,billy) $ because $  BB(t_1+ d_b)$
holds after Suzanne's throw (12).  Billy's stone cannot hit the
bottle, because it is already broken when his stone could hit it
and we have just $[ T(t_1,d_b,billy)]\top$ ((6), Billy has
thrown).

\item Billy's stone hits the bottle, which breaks, \textbf{before}
Suzanne's stone could possibly hit the bottle.

\[\begin{array}{ll}
  (13) & t_1 + d_b +d_1 < d_s \\
  (14) & \Box (BB( t_1 + d_b +d_1) \rightarrow BB(d_s)) \mbox{  from (3) and (13) }\\
  (15) & \neg BB(t_1 + d_b)  \mbox{  by persistency from (4), see (9)}\\
  (16) & [ T(t_1,d_b,billy)] H(t_1 + d_b,billy) \mbox{  from (1) and (15) }\\
  (17) & [T(t_1,d_b,billy)]BB(t_1 + d_b + d_1) \mbox{  from (2), (16), K  and   (A2) }\\
  (18) & [T(t_1,d_b,billy)]BB(d_s) \mbox{  from (14), (17), K and   (A2) }

\end{array}\]

Here, Suzanne's stone, which could hit the bottle  at instance
$d_s$, will  not hit it since we have $BB(d_s)$ and therefore  the
precondition $ \neg BB(d_s)$ is not more true.  The law $\Box(
\neg BB(d_s) \rightarrow [ T(0,d_s,suzy)] H(d_s,suzy)) $ cannot be
used to derive $[ T(0,d_s,suzy)] H(d_s,suzy) $ because $ BB(t_1+
d_b + d_1)$ holds after Billy's throw (17).  All    we have is
$[T(0,d_s,suzy)]\top$ (Suzanne throws).

 \item Suzanne's and Billy's stone hit the bottle precisely at the same moment.
\[\begin{array}{ll}
  (19) & t_1 + d_b  = d_s \\
  (20) & \neg BB(t_1 + d_b)  \wedge \neg BB(d_s) \mbox{  by persistency from (4), see (9)} \\
  (21) & [ T(0,d_s,suzy)] H( d_s,suzy) \mbox{  like (10) }\\
  (22) &   [ T(t_1,d_b,billy)] H(t_1 + d_b,billy) \mbox{ as (16) }\\
  (23) & [T(0,d_s,suzy)]BB(d_s + d_1) \mbox{ from (21)}\\
  (24) & [T(t_1,d_b,billy)]BB(t_1 + d_sb + d_1) \mbox{ from (22)) }

\end{array}\]

In this case, both stones hit the bottle which breaks as a result
of Suzanne's throw and Billy's throw.

  \ee

\section{Conclusion and Related Work}
Modal logic approaches to action theories define a space of states
but cannot handle time, neither explicitly not implicitly
\cite{GiordanoMartelliSchwind:99,CastilhoGasquetHerzig:99}. In
situation calculus  \cite{Reiter:01,McCarthy:02}
 reasoning about time was not foreseen, properties change discretely and actions
 do not have durations. Remember that in situation calculus, there is a starting state,
 $s_0$ and    for any action $a$ and state $s$, $do(a,s)$ is a
 resulting state of $s$. One can consider that the
 set of states is given by $\{s: \exists a_1 \dots a_n  s=do(a_n,
 do(a_{n-1}, \dots do(a_1, s_0)))\}$. Hence the temporal
 structure of situation calculus is discrete and branching and does not allow
 for actions of different duration neither for preconditions or
 results which become true during the action execution or later
 after the action is ended.

Javier Pinto has extended  situation calculus in order to
integrate time \cite{Pinto:98b}. He conserves the framework of
situation calculus and introduces a notion of time. Intuitively,
every situation $s$ has a starting time and an ending time, where
$end(s,a) = start(do(a,s))$ meaning that situation $s$ ends when
the succeeding situation $do(a,s)$ is reached. The end of the
situation $s$ is the same time point as the beginning of the next
situation resulting from the occurrence of action $a$ in $s$. The
obvious asymmetry of the $start$ and $end$ functions is due to the
fact that the situation space has the form of a tree whose root is
the beginning state $s_0$. Thus, every state has a unique
preceding state but eventually more that one succeeding state.

Paolo Tereziani proposes in \cite{Terenziani:02} a system that can
handle  temporal constraints between events  and temporal
constraints between instances of events.

In this present article, we have introduced a new modal logic
formalism which can handle simultaneously states and time. We did
not address here the problem of the persistency of facts over time
(or over the execution of actions), because we wanted to focus on
the modal temporal formalism. We have adopted a solution similar
to the one presented in \cite{Schwind:98}, i.e. ``weak'' frame
laws are nonmonotonically added to the theory. But this solution
is a bit more complicated in the case of our first-order action
logic presented in this paper, because we need to restrict
ourselves to a decidable subset of \FOMOL.

Concerning the implementation, we  use a labelled analytic
tableaux approach including an abductive mechanism for the weak
persistency laws, which will be described in more detail in a
following paper.

We will apply this formalism to planning problems where a hybrid
approach (states and time) can be very powerful. The idea is to
infer temporal constraints from a \FOMOL specification in order to
create a plan for a problem

\end{sloppypar}



\bibliographystyle{plain}

\appendix
\section{Appendix: Proofs of the theorems}

\subsection{Soundness and Completeness}
Soundness is easy and left tyo the reader. For completeness, 
in this part we will show that every \FOMOL - valid formula is a
theorem of \FOMOL. Our proof  is along the same lines as
\cite{Schuette:78}. Subsequently, formulas   are always grounded.
\begin{sloppypar}

\begin{de}\label{de:inconsistent}
A set of formulas, $s$, is called \FOMOL -inconsistent if it
contains a finite subset $\Xnset{\phi}{k}$ with
$\Xnforl{\phi}{k}{\FOMOL}$. Otherwise $s$ is called \FOMOL -
consistent (or consistent).
\end{de}
\end{sloppypar}

 Let $s$ be a set of formulas. Then we denote by
$V_{t}(s)$ the set of all terms occurring in formulas of $s$. We
denote by $P(s)$ the set of all \FOMOL - formulas   containing
only nonlogical symbols (terms, action terms and predicate
symbols) of formulas of $s$.

\begin{de}\label{de:complete}
A set of formulas $s$ is called \mbox{complete} (or \FOMOL -
complete) if
\begin{enumerate}
\item $s$ is consistent. \item $s$ is maximal, i.e. for all $\phi$
in $P(s)$ holds: if $\phi \not\in s$ then $s \cup  \{\phi\}$ is
inconsistent. \item $s$ is  saturated, i.e. for every existential
formula $\exists x \phi~\in~s$ where $x$ is a  variable, there is
a formula $\phi_{c}^{x} \in s$  for some constant $c \in
V_{t}(s)$.
\end{enumerate}
\end{de}

\begin{lem}\label{Lindenbaum}
Every consistent set of formulas can be extended to a complete set
of formulas.

\begin{proof}
Let $s$ be a consistent set of formulas and let $c_1$, $c_2$,
$\ldots$ be a sequence of `new' object variables not in
$V_{t}(s)$. We define $P^\star(s) = P(s) \cup \{\psi: \psi$ is a
formula with variables from $c_1,c_2  \ldots \}$. Let $\{\exists
x_i \phi_i(x_i) \}_{i=1,\ldots}$ be an enumeration of {\em all}
existential formulas of P*(s). Then we form a new set of formulas
$s^\sim$, by adding to $s$ all formulas $\exists x_i
\phi_i(x_i)\rightarrow \phi(a_i), i = 1, 2, \ldots$ for every
existential
formula of $P(s)$. It is easy to see that $s^\sim$ is consistent.\\
  $s^\sim$ is extended to a complete (maximal) set of formulas  $s^\star$ as follows: Let $\Theta = \{s': s \subset s'$ and $s'$ consistent\}. Let $H$ be a chain in
$ \Theta  $, i.e. $H \subset  \Theta  $  and if $s_1$, $s_2 \in H$
then $s_1 \subset s_2$ or $s_2 \subset s_1$. Then $\bigcup H$ is
an upper bound of $H$ in  $ \Theta  $, since $s \subset \bigcup H$
for all $s \in H$. It is easy to see that $\bigcup H$ is
consistent. Therefore  $\bigcup H \in \Theta$ and, by Zorn's
Lemma, $\Theta$ then has a maximal element,  $s^\star$. It is easy
to see that $s^\star$ is complete.

\end{proof}

\end{lem}

The following properties of complete sets are straightforward.
\begin{lem}\label{lemcomplete}
Let $s$ be a complete set of formulas. Then
\begin{enumerate}
\item  $\phi \in s$ if and only if   $\neg   \phi \not\in  s$

\item If $\phi \in s$ and $\psi \in P(s)$ and $\vdash_{\FOMOL}
\phi \rightarrow  \psi$   then  $\psi \in s$
\item  If $\Box\phi \in s$   then  $[a]\phi \in s$

\item If $\phi \vee \psi \in s$   then   $\phi \in s$ or $\psi \in
s$

\item If $\phi \vee \psi \in P(s)$   and   $\phi \in s$ or $\psi
\in s$ then  $\phi \vee \psi \in s$

\item If  $c \in V_{t}(s)$, then $\phi^{x}_{c} \in s$   if and
only if   $\exists  x \phi \in s$

\item   $t=t \in s$  for any term $t\in V_{t}(s)$

\item   If $t=t' \in s$  then $t'=t \in s$

\item  If $t=t' \in s$ and  $t'=t'' \in s$ then  $t=t'' \in s$

\end{enumerate}

The last three items show that  $t=t' \in s$ defines an
equivalence relation $\sim_{s}$ over $V_{t}(s)$, where $t \sim_{s}
t'$ iff $t=t' \in s$. The equivalence class of $t$ according to
$\sim_{s}$ is denoted by $[t]_s$.

\end{lem}

\begin{de}\label{desa}
Let $s$ be a set of formulas  and $a \in \doppel{A}_{t}$. Then we
define
\begin{enumerate}
\item     \mbox{$s^a  = \{\phi : [a]\phi \in s\}$}
\item  \mbox{$s^{\Box}  = \{\phi :  \Box \phi \in s\}$}

\end{enumerate}

    \end{de}

    \begin{lem}\label{lemsa}
\begin{enumerate}
\item     $s^a$ is inconsistent  iff  $[a] \bot \in s$
\item $s^{\Box} \subseteq s^a$

\end{enumerate}
 
     \end{lem}

\begin{lem}\label{terms}
Let $s$ be a complete set of formulas and $\phi$ a formula such that  $s^a \cup \{\phi\}$ is 
consistent. Let  $s'=(s^a \cup \{\phi\})^*$  the complete
extension of $s^a \cup \{\phi\}$ (which exists by lemma \ref{Lindenbaum}). Then
$V_t(s) = V_t(s')$
\begin{proof}
Since $s$ is complete, it is saturated, hence for every existential formula $\exists x \phi \in s$ there is a formula ${\phi^x}_c \in s$ for some $c\in V_t(s)$. The crucial point concerns existential formulas in $s^a$ not in $s$. But for those formulas we have that  $\Box \exists x \phi \in s$ and therefore $\exists x \phi \in s^a$, by $T$. Hence existential formulas in $s^a$ are also existential formulas in $s$. But this means that all terms in s' are terma in $s$.
\end{proof}
\end{lem}


\begin{de}\label{desyscomplete}
 A set $\mathcal S$ of sets of formulas is called
{\em complete} if\\
1. Every element $s$ of $S$ is a complete set of
formulas.\\
2. For all  $s \in S$ and $\phi \in P(s)$, if $s^a \cup \{\phi\}$
is consistent, then there is $s'\in S$ such that $s^a \cup
\{\phi\} \subset s'$
\end{de}

\begin{re}
If $[a]\bot \in s$, then, since  $s^a$ is not consistent, it is  and not contained
in any $s' \in S$.
\end{re}

\noindent Now we construct the canonical model starting from a complete set
of formulas,  $s$.
\begin{sloppypar}

\begin{lem}\label{lemextcomsys}
For every complete set of formulas $s$, there is a
\FOMOL-structure \DYS, such that $s \in \mathcal{W}$.

\begin{proof} : \\
For every $\phi \in P(s)$ and every $a \in \doppel{A}_t$ such that $s^a \cup \{\phi\}$ is
consistent, we extend $s^a \cup \{\phi\}$ to a complete set of
formulas $s'$, which exists by lemma \ref{Lindenbaum}. So we form
successively sets of formula sets $S_0$,
$S_1$,$\ldots$ as follows\\
$S_0 : = \{s\}$ \\
$S_{i+1} := \{X: X$ is a complete
extension of $Y^a \cup  \{\phi\}$ where $Y \in S_i$ and $\phi \in P(s)$ and  $Y^a \cup  \{\phi\}$ is consistent\}\\
${\mathcal W}: = \bigcup\{S_i: i \in \omega\}$.\\

Because of remark \ref{terms}, $V_{t}(s)=V_{t}(s')$ for $s, s' \in {\mathcal{W}}$. We note $V_{t}(s)=V_{t}$, $[t]_s = [t]$ and $\sim_s = \sim$. \\

For every $w \in  {\mathcal{W}}$, ${\mathcal{S}}_{w} =
({\mathcal{O}}, {\mathcal F}_{w}, {\mathcal{P}}_{w})$ is a
classical structure where the object set ${\mathcal{O}}$ is
defined as the set of $\sim$-equivalence classes over $V_{t}$. \bi
\item ${\mathcal{O}} {=}_{def} \{[t]: t\in  V_{t}\}$

\item For every $t\in {\mathcal{O}}$,  $\tau_0(w,t) = [t]$

\item for every n-ary function symbol $F\in \doppel{F}$ and for
every tuple of terms  $t_1, \dots t_n \in V_{t}$,
$\tau_{1}(w,F)([t_1], \dots [t_n]) = [F(t_1, \dots t_n)]$

\item ${\mathcal{P}}_{w} $ is the set of predicates
$P^{w}$, where for every predicate symbol $P$ of arity
$n=\Card{P}$, $P^{w}$ is the predicate defined by  $(\tau_1(t_1),
\ldots \tau_1(t_n)) \in P^{w} \mbox{ if and only if } P(t_1,
\ldots t_n) \in w$. We set $\tau_2(w,P) = P^{w}$

\item $\mathcal{A}$ is the set of action functions $a$.  For every
action symbol $a$ of arity $\Card{a} = m$, and terms $\Xnseq{t}{m}$,
$a(w,\Xnseq{t}{m}) \subseteq {\mathcal{W}}$ is defined by $w' \in
a(w, \Xnseq{t}{m})  \mbox{ if and only if }  w^{\Xnterm{a}{t}{m}}
\subseteq w'$

\item We set $\tau_3(a) = a$ \ei

\end{proof}
\end{lem}

We show that 
$\MM$ is a \FOMOL-structure.  By the construction,
$\WW$ is a complete set of sets.

\begin{lem}
 $ \tau_3(a)(w, \tau_{0}(t_1), \ldots, \tau_{0}(t_{n})) \subseteq   R(w)$.
 Let be $w' \in  \tau_3(a)(w, \tau_{0}(t_1), \ldots, \tau_{0}(t_{n}))$. Then $w^{\Xnterm{a}{t}{m}}
\subseteq w'$ by construction of the model. But then $w^{\Box} \subseteq w'$  by lemma \ref{lemsa}, from which follows that $w'\in R(w)$. 
\end{lem}

Truth value $\tau(w,\phi)$ for formula $\phi$ and world $w$ is
defined by induction over the  construction of formulas  as usual.

In order to show that $\tau(s,\phi) = t$ if and only if $\phi \in
s$, we need the following lemma:

\begin{lem}\label{settruth}
Let \DYS be the \FOMOL-structure constructed from a complete
formula set $w$ according to lemma \ref{lemextcomsys} and
 $w \in W$. \\
Then for every action symbol $a$ of arity $n$,
 $[a]\phi \in w$ {\em if and only if} \\
{\em for every tuple of terms }  $\Xnseq{t}{n}$ and $w' \in \WW$,
{\em if}  $w' \in \tau_3(a)(w,\Xnseq{t}{n}$, {\em then} $\phi \in
w'$.

\begin{proof}\\
If $\phi$ is $\bot$ then this lemma trivially holds because then $\tau_3(a)(w,\Xnseq{t}{n} $ is empty.\\
Let $\phi$ be different from  false.\\
$(\Rightarrow)$: if $[a]\phi \in w$ then $\phi \in w^a$ and therefore for every $w' \in a(w,\Xnseq{t}{n})$,  $\phi \in w'$ by the definition of $\MM$.\\
$(\Leftarrow)$: Let be $\Card{a} =n$. Consider  $w^a$. We first
show that $w^a \cup \{\neg
\phi\}$ is inconsistent. \\
Assume for the contrary, that  $w^a \cup \{\neg \phi\}$ is
consistent. Since $\mathcal{W}$ is a complete set of formula sets,
according to definition \ref{desyscomplete}, there is $w' \in \WW$
and $w^a \cup \{\neg \phi\}\subseteq w'$. From this follows $w^a
\subseteq w'$, hence
$w' \in a(w,\Xnterm{t}{\Card{a}})$, by the constuction of \MM in lemma \ref{lemextcomsys}. Since $\neg \phi \in w'$ and $w'$ is complete, $\phi \not\in w'$, which contradicts the hypothesis. \\
Therefore, $w^a \cup \{\neg \phi\}$ is inconsistent. Therefore
there exist formulas \mbox{$\Xnseq{\phi}{k} \in w^a$} such that
$\vdash \phi_1 \wedge \ldots \wedge \phi_k \rightarrow \phi$ from
which $\vdash [a](\phi_1 \wedge \ldots \wedge \phi_k) \rightarrow
[a]\phi$ by rule 1 of the logic \FOMOL and therefore $\vdash
[a]\phi_1 \wedge \ldots \wedge [a]\phi_k \rightarrow [a]\phi$. But
$\Xnseq{[a]\phi}{k} \in w$, and therefore $[a]\phi \in w$, by the
lemma \ref{lemcomplete}. 2.
\end{proof}
\end{lem}

The next lemma concludes our proof.

\begin{lem}\label{lemtopin}
For every closed formula $\phi \in P(s)$, $$\tau(w,\phi) = t
\mbox{ if and only if }   \phi \in w$$ The proof is
straightforward, by induction over formulas using the lemmata
\ref{lemcomplete} and \ref{settruth}.
\end{lem}

Proof of the completeness theorem:\\

\begin{proof} Assume for the contrary
that there is a \FOMOL valid formula $\phi$ which is not deducible
in \FOMOL. Then  $\{\neg \phi\}$ is a  consistent set of formulas,
which can be extended to a complete set of formulae $s$, by Lemma
\ref{Lindenbaum}, where $\neg \phi \in s$. By Lemma
\ref{lemextcomsys}, there is a \FOMOL-structure \DYS where \WW is
a complete system of sets and $s \in \WW$. By Lemma
\ref{lemtopin}, $\tau(s, \neg \phi) = t$ and hence $T(s,\phi) =
f$,  which contradicts the validity of $\phi$.
\end{proof}

\end{sloppypar}

\end{document}